\documentclass[10pt,a4paper]{article}

\usepackage[utf8]{inputenc}
\usepackage[T1]{fontenc}
\usepackage{lmodern}
\usepackage{microtype}
\usepackage{graphicx}
\usepackage{booktabs}
\usepackage{amsmath}
\usepackage{amssymb}
\usepackage{siunitx}
\usepackage{xcolor}
\usepackage[round,authoryear]{natbib}
\usepackage[hidelinks]{hyperref}
\usepackage[margin=0.95in]{geometry}
\usepackage{caption}
\usepackage{subcaption}
\usepackage{multirow}
\usepackage{indentfirst}

\newcommand{\errkm}{\ensuremath{\mathrm{err}_{\mathrm{km}}}}
\newcommand{\refkm}{\ensuremath{d_{\mathrm{ref}}}}
\newcommand{\Ldist}{\ensuremath{L_{\mathrm{dist}}}}

\title{\textsc{GeoContext}: One Context Ladder, Two Failure Modes\\
in Vision--Language Geolocation\\
\large Flat Reliance on User-Provided Location Context,\\
and False Confirmation of Location Claims}

\author{
  Yifan Zhang\\
  \small \texttt{powerfan233@gmail.com}
  \and
  Kai Wang\\
  \small \texttt{wangjinjie722@gmail.com}
}

\date{\today}

\begin{document}
\maketitle

\begin{abstract}
Visual geolocation benchmarks ask a model where a photograph was taken and score
how far off it is. Real users rarely ask that in isolation: they volunteer where
they think they are --- ``I took this near $X$'' --- and that sentence is usually
\emph{true but imprecise}. How a model should weigh such a hint is unmeasured,
and the same weighing governs a deployed decision: whether a photograph confirms
a location \emph{claim}, as when a platform asks a driver to photograph a
drop-off.

We present \textsc{GeoContext}, one resource supporting two tasks:
\textsc{GeoHint}, open-ended localization given a true but coarse hint, and
\textsc{GeoVerify}, binary verification that an image was taken within
\SI{150}{\metre} of a claimed place, scored with signal detection theory. The
shared resource is a \emph{context ladder} that stratifies mined reference
points by distance ($0.5$--$6$\,\si{\kilo\metre}) and by
\emph{referenceability} --- how often people use a place to say where they are,
which we show is distinct from fame --- so that the image is held fixed while
only the hint varies. The
release covers \num{109} sites in \num{30} cities, \num{5} VLMs, and
\num{21933} scored \textsc{GeoHint} and \num{6270} \textsc{GeoVerify}
responses.

Our evaluation yields three main findings. Reliance on the hint is flat along both axes --- \num{1.5}
points across referenceability, under \num{3} across distance --- while the
error that relying produces grows steadily, so the cost scales and the reliance
does not. Where the imagery is illegible the median ratio of error to hint
distance is $1.00$: the answer \emph{is} the hint. Where it is legible that
ratio falls to \numrange{0.24}{0.69}, and one model of five is measurably less
accurate with a hint than without one. On \textsc{GeoVerify} no model reaches
$d'=1$ at the tolerance a deployment needs, ranking by accuracy inverts the
ordering because models differ sharply in response bias, and
\SI{83.8}{\percent} of false acceptances are stated at confidence
$\geq\num{0.8}$. Code, data and every audit decision are released.
\end{abstract}

\section{Introduction}

Visual geolocation benchmarks score a model by how far its predicted coordinate
falls from the truth. That framing hides a failure mode that ordinary users hit
constantly. Consider the interaction that motivated this work: one of us
photographed a building while traveling, asked a VLM what it was, and received
the name of a \emph{different, more famous} building a few hundred meters away.
The region was right. The entity was wrong. A coordinate-distance metric barely
registers the error; a user trying to identify what is in front of them is
completely misled.

Such interactions are rarely context-free. Users volunteer what they know: ``I
took this near Bastille'', ``somewhere in Shoreditch''. These statements are
\emph{true but under-specified}---they constrain the answer to a neighborhood,
not a building. Prior work on textual context in geolocation has studied the two
extremes: adversarial text \emph{inside} the image
\citep{signpost}, and out-of-image captions that either give the answer away or
are irrelevant \citep{koreageo}. The middle case---context that is honest,
helpful in principle, and merely coarse---is where real users live, and it is
unstudied.

We ask: \textbf{does a VLM's reliance on user-provided location context track
how informative that context is?} We find it does not discriminate along
either axis, and the two readings differ. How commonly a place is used to give
directions can be judged from its name, so a model that repeats an obscure
reference point as readily as a universally known one is failing at something
it could have done. How far away that place is cannot be read off the name, so
flatness there is the absence of a signal rather than a failure to use one ---
but the error that repeating it produces grows with that distance regardless,
which is what makes the flatness consequential.

The same failure has a second face, with a deployed decision attached to it.
Turn the question around --- instead of asking the model to \emph{produce} a
location given a hint, ask it to \emph{check} one: the photo is claimed to have
been taken at a particular place; is that true? This is the shape of visual
arrival verification, and a model that cannot discount an unreliable hint when
naming a place is unlikely to reject a false claim when checking one. We
therefore build both tasks on one resource and report them together.

\paragraph{Contributions.}
\begin{enumerate}
  \item \textbf{A resource supporting two tasks.} A context ladder, mined from
  Wikidata and rated by a text-only auditor for referenceability, which
  stratifies reference points by distance $\times$ referenceability and
  transfers to a new city with no human labor (Sec.~\ref{sec:ladder}). It
  supports both \textsc{GeoHint} (Sec.~\ref{sec:geohint}) and, reusing the same
  coordinates as decoys, \textsc{GeoVerify} (Sec.~\ref{sec:geoverify}).

  \item \textbf{Reliance that stays flat while its cost grows.} Models repeat a
  hint almost as often when it is distant and obscure as when it is near and
  universally known, while the error that repeating it produces grows steadily
  with its distance (Sec.~\ref{sec:mechanism}). Referenceability is legible
  from the name and is still ignored; distance is not, and the models discount
  it only where the image lets them check it. The cost of relying on the hint
  scales; the reliance does not.

  \item \textbf{A consequence conditional on image legibility, and a mitigation
  that follows the same mechanism.} Where the imagery is illegible the answer
  collapses onto the hint (median error/distance ratio $1.00$); where it is
  legible the image does the work instead, and one of the five models is
  reliably less accurate for having been given true information
  (Sec.~\ref{sec:baseline}). Warning the model that the user may be wrong is
  obeyed in both regimes but helps only in the legible one
  (Sec.~\ref{sec:mitigation}); a structured evidence checklist helps in neither,
  against pre-registered criteria (Sec.~\ref{sec:chain}).

  \item \textbf{Verification scored without a geocoder, and the measurement
  hazards of scoring with one.} \textsc{GeoVerify}'s ground truth is a haversine
  comparison, so no geocoder, judge model or output threshold enters the
  measurement; we score it with signal detection theory and summarize each model
  by a discrimination radius (Sec.~\ref{sec:geoverify}). By contrast we document
  four faults in the geocoded metric, one of which reversed our own headline
  result, as a general hazard for distance-thresholded geolocation scoring
  (Sec.~\ref{sec:scoring}).
\end{enumerate}

\section{Related Work}

\paragraph{Image-only geolocation.} The dominant line predicts coordinates from
the image alone \citep{translocator}, evaluated with distance-threshold accuracy
$\mathrm{Acc}@k\,\si{\kilo\metre}$. Because test images are drawn near-uniformly
over the globe, most of them are wilderness, road, or low-density periphery ---
not where people photograph things and ask what they are. Recent LLM-era
benchmarks inherit both the sampling and the metric \citep{imageo}. Under a
distance metric, naming the building next door is an \SI{80}{\metre} error and
therefore indistinguishable from a perfect answer; the failure this paper is
about is invisible by construction. We keep a distance threshold for
\emph{task performance}, because it is the measure that is comparable across
sites, and carry the entity-level question on two other measures instead: how
often the answer is the hint repeated back (echo rate), and how far the answer
lands relative to how far away the hint was
(Sec.~\ref{sec:reliance}).

\paragraph{Text alongside the image.}
SIGNPOST-Bench \citep{signpost} renders adversarial place names \emph{into} the
image; predictions shift by hundreds of kilometers. KoreaGEO \citep{koreageo}
pairs images with out-of-image captions that are either irrelevant or directly
revealing. Both manipulate text that either contradicts the image or gives the
answer away. Ours does neither: it is true, out-of-image, and coarse, so failures
can appear only at the \emph{entity} level while the region stays correct.

\paragraph{Anchoring on salient objects.} HoloGeo \citep{hologeo} reports that
models over-weight visually salient landmarks and proposes two bias metrics.
That work concerns what is \emph{inside} the image; we concern the text the user
adds beside it. Their metrics also require output probability distributions and
so cannot be computed for API-only models --- a reproducibility constraint we
deliberately avoid.

\paragraph{Auxiliary hints, and why they have not been studied as a variable.}
EarthWhere \citep{earthwhere} is the closest prior work: its WhereStreet split
supplies each item with a free-text \texttt{HINT} taken from the source video
(e.g.\ ``this image was taken on my way to school''). Two features of that
design leave the gap we fill. First, the hint is whatever the narrator happened
to say: its distance from the target and its usefulness are neither controlled
nor recorded, and there is no matched no-hint condition, so no dose--response
curve can be estimated. Second, the scoring protocol explicitly \emph{removes}
the behavior we measure --- credit is computed one level below whatever the hint
names, on the instruction to ``treat the Hint as free information; exclude it
from credit'', with a further rule zeroing predictions that merely copy it.
Guarding against the behavior concedes that it happens; it also converts it from
a phenomenon into a scoring problem. We make it the dependent variable instead,
and vary the hint's distance and referenceability so that the reliance a model
shows can be compared against the reliance the hint warrants.

\paragraph{Verifying a location claim.} A second line of work asks not
\emph{where} an image was taken but whether a \emph{claimed} time or place is
consistent with it. \citet{imageguard} infer the sun position from a single
shadow and compare it against the position implied by the claimed timestamp and
coordinates, correctly flagging \SI{91.5}{\percent} of falsified photographs;
\citet{timestamp} learn the same consistency check end-to-end, raising accuracy
on timestamp manipulation from \SI{59.0}{\percent} to \SI{81.1}{\percent}. Both
verify a claim through a physical channel --- shadow geometry, illumination ---
that is independent of what the scene depicts. At industrial scale,
\citet{dumapper} verify points of interest by matching signboard imagery against
an archived database, replacing crowdworker effort with retrieval. The question
we ask is whether a general-purpose VLM can carry this decision through the
\emph{semantic} channel instead, judging from the scene itself whether a named
place is where the photograph was taken, at the tolerance a deployment needs.
The nearest benchmark in spirit is \citet{geoxbench}, which pairs panoramas with
satellite imagery and asks models for relative pose; its supervision is
cross-view geometry rather than a named claim a user might assert, and it does
not measure the false-acceptance behavior that matters when the claim is
adversarial.

\section{The \textsc{GeoContext} resource}
\label{sec:resource}

Everything in this section is shared by both tasks. We call the resource
\textsc{GeoContext}: a programmatic pipeline that, for any city, produces
photo sites, street-level imagery of those sites, and a \emph{context ladder}
of nearby reference points stratified by distance and referenceability. Two
evaluation protocols run on top of it:

\begin{description}
  \item[\textsc{GeoHint}] (Sec.~\ref{sec:scoring}, \ref{sec:results-hint}) ---
  open-ended naming. The model sees the image plus one true but coarse sentence
  of user-supplied context, and must name where the photo was taken. This is
  the task an ordinary user performs when they ask a chatbot what they are
  looking at.
  \item[\textsc{GeoVerify}] (Sec.~\ref{sec:geoverify}, \ref{sec:results-verify}) --- claim verification.
  The model sees the image plus a \emph{claimed} location and must answer
  whether the photo was taken within a stated tolerance of that claim. This is
  the task a platform performs when it asks a driver for a photo to confirm a
  drop-off.
\end{description}

The two tasks read the same sites, the same images and the same audited
reference-point pool; they differ only in what is asked and how the answer is
scored. \textsc{GeoHint} needs a geocoder, and its threshold is applied to the
geocoded output. \textsc{GeoVerify} needs no geocoder, and its tolerance
$\tau$ is applied to two known coordinates rather than to anything the model
produced, so the ground truth is a deterministic distance comparison
(Sec.~\ref{sec:geoverify-metric}).

\begin{figure}[t]
\centering
\includegraphics[width=\textwidth]{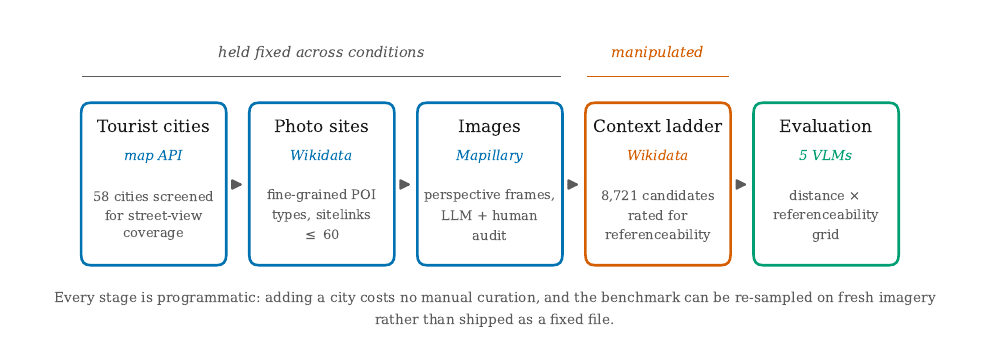}
\caption{Construction pipeline. City screening, site selection, imagery
retrieval and reference-point rating are all programmatic; the only human step
is a spot audit of the retrieved images (Sec.~\ref{sec:scope}). The first three
stages are held fixed within a site, so the reference point is the sole
manipulated input. Because the pipeline is a sampler rather than a fixed file,
the benchmark can be regenerated on imagery captured after a given date --- the
route we intend to use for contamination control.}
\label{fig:pipeline}
\end{figure}

\subsection{Scope: dense, walkable urban cores}
\label{sec:scope}

Figure~\ref{fig:pipeline} summarises the construction.
Distance bands only mean the same thing across cities if the cities have
comparable spatial granularity. We make the inclusion criterion explicit and
measurable. For a site, let \Ldist{} be the median nearest-neighbor distance
among OSM \texttt{place=neighborhood|suburb|quarter} \emph{point} nodes within
\SI{6}{\kilo\metre}. We use point nodes rather than administrative polygons
because polygon coverage is wildly uneven---New York returns \num{11} polygons,
and informal but universally used names (SoHo, Nolita, Tribeca) have none.

Two details matter.
\emph{First}, addressing units must be excluded. Of \num{800} ``neighbourhoods''
within \SI{6}{\kilo\metre} of Shibuya, \num{573} (\SI{72}{\percent}) are
\textit{X-chōme} street-numbering blocks. Including them yields
$\Ldist = \SI{0.24}{\kilo\metre}$ for Tokyo versus \SI{0.51}{\kilo\metre} for
Paris, supporting a conclusion---``Tokyo's neighbourhoods are twice as
fine-grained''---that is purely an artifact of Japanese OSM tagging convention.
After exclusion Tokyo's three sites give $0.34/0.44/0.58$, in line with
everywhere else.
\emph{Second}, the measure captures granularity, not motorised
accessibility (Sec.~\ref{sec:limits}).

For the \num{19} hand-picked sites, measured individually,
$\Ldist \in [0.33, 0.65]\,\si{\kilo\metre}$, a spread of only $1.98\times$; we
therefore use \emph{fixed} absolute distance bands. San Jose, a car-oriented
city we measured but excluded, gives $\Ldist = \SI{0.99}{\kilo\metre}$ with only
\num{21} neighborhood nodes in the same radius. \textbf{Car-oriented
low-density urban areas are out of scope}; there ``nearby'' is set by drive
time, not neighborhood granularity, and would require a different band design.

For the \num{90} algorithmically selected sites the criterion was applied at
\emph{city} level --- one representative center coordinate per city, all
\num{30} within $[0.32, 0.65]\,\si{\kilo\metre}$ --- rather than per site. Two
cities, Tokyo and Bogot\'a, sit at \num{0.328} and \num{0.327} and were admitted
by lowering the floor from \num{0.33} to \num{0.32}; Seville at \num{0.316} was
not. Per-site verification for this batch is outstanding and is listed in
Sec.~\ref{sec:limits}.

\subsection{Sites and images}

\num{109} sites in \num{30} cities, from two selection procedures. \num{19}
were hand-picked across \num{6} countries: Paris (Montmartre, Bastille,
\^Ile de la Cit\'e, Canal Saint-Martin, Marais), London (Shoreditch, Covent
Garden, Notting Hill), Barcelona (Gr\`acia, El Born), San Francisco (Mission,
North Beach, Hayes Valley), Mexico City (Roma Norte, Condesa), Tokyo (Shibuya,
Shimokitazawa, Yanaka), New York (SoHo). The remaining \num{90} were selected
automatically, three per city across \num{30} cities, by the procedure of
Sec.~\ref{sec:site-selection}. The hand-picked cities are a subset of the
\num{30}, so those cities carry more sites than the rest.

\paragraph{Site selection deliberately avoids the most iconic landmarks}
(Eiffel Tower, Sagrada Família, Zócalo). Two reasons: iconic targets saturate
the baseline and lose discriminative power; and the failure mode under study
\emph{is} substituting the nearby iconic landmark, which is meaningless if the
target is that landmark. For the hand-picked sites this was a judgement call by
the authors; the automatic procedure below replaces that judgement with a
reproducible rule, and the two sets should be read as offering different
guarantees.

\subsubsection{Automatic site selection}
\label{sec:site-selection}

Candidate cities are a manually compiled list of \num{58} tourist destinations
(the union of several published rankings; the list is an \emph{input} to the
pipeline and can be edited freely, since the two automatic filters below do the
work). A city is admitted if it has $\geq \num{30}$ perspective Mapillary images
within \SI{1}{\kilo\metre} of its center and its \Ldist{} falls in
$[0.32, 0.65]\,\si{\kilo\metre}$.

Within an admitted city we query Wikidata for entities within
\SI{3}{\kilo\metre} of the center that (i) instantiate one of \num{16}
photographable types --- museum, park, church, theatre, market, bridge, square,
library, castle and similar, deliberately excluding administrative divisions,
which are areas rather than objects --- and (ii) have at most \num{60}
Wikipedia sitelinks, a ceiling that removes only the extreme top of the fame
distribution. Candidates are then \textbf{shuffled} and checked in random order
for street-view coverage ($\geq \num{8}$ perspective images within
\SI{150}{\metre}), taking the first three that pass and skipping any within
\SI{250}{\metre} of an already-chosen site.

The shuffle is the detail that matters. Querying Wikidata returns candidates
ordered by sitelinks, and an earlier version simply walked that list until it
had three with coverage --- which is equivalent to ``the most famous eligible
landmarks in this city'' and produced a test batch consisting almost entirely of
tourist-postcard subjects (Anne Frank House, Brandenburg Gate). Every one of
them was below the sitelinks ceiling, so the fault was in the traversal order,
not the threshold. We note also that a narrower sitelinks \emph{band} does not
work as a proxy for referenceability. Across the \num{8649} audited candidates
for which Wikidata reports a sitelink count, the range $5$--$40$ holds
\SI{76.7}{\percent} of the places rated highly referenceable, missing close to a
quarter of them, and internally it spans every referenceability level from
$0$ to $5$. Sitelinks can exclude extremes; they cannot select the middle.

\paragraph{Replication unit is the site, not the image.} Multiple images at one
site are near-duplicates---same buildings, same street, correlated errors---
whereas the generalization claim is ``does this hold somewhere else''. We
therefore prefer more sites over more images per site.

\paragraph{Images} come from Mapillary (CC-BY-SA); we release image IDs and
coordinates rather than pixels. Candidates pass an automatic daylight/brightness
filter and are then audited twice, \emph{independently}: by a human, and by a
vision model (\texttt{deepseek-vision}, which is not among the evaluated
models). The model rates each image on answer leakage --- text burned into the
frame naming a place or printing coordinates, the one hard exclusion --- as well
as scene type, shooting through glass, visible tourist infrastructure, and
overall quality (prompt in Appendix~\ref{app:prompt-image}). The human marks each image keep or exclude with a free-text
reason. Neither reviewer sees the other's verdict before deciding, so agreement
between them is meaningful. Of \num{429} audited candidates (\num{170}
hand-picked, \num{259} automatic) the human kept \num{347}; \num{159} of these
enter the evaluation (\num{59} hand-picked, \num{100} automatic), sampled to
cover every site.

\paragraph{The two audits are not redundant, and agreement statistics say so.}
Human and model agree on \num{343} of \num{429} images (\SI{80.0}{\percent}),
but Cohen's $\kappa = \num{0.185}$: the high raw agreement comes mostly from the
large majority of images that are unproblematic for both, and the two judgements
are close to independent where it matters. The disagreements are systematic
rather than noisy, and each direction has a cause:

\begin{itemize}
  \item \textbf{Human excludes, model accepts} (\num{65} images). The human's
  operative criterion --- ``nothing in this frame identifies a location'' ---
  is not among the model's fields at all. A blank wall photographed sharply in
  good light scores well on every dimension the model rates.
  \item \textbf{Model rejects, human keeps} (\num{21} images), of which
  \num{17} are shots from inside a vehicle. The model treats
  \texttt{scene\_type} $\neq$ \texttt{street} as disqualifying; the human keeps
  the frame whenever the street outside is legible through the windscreen.
\end{itemize}

Answer leakage is the category that most rewards running both tracks, because
neither catches all of it. Nine images carry GPS coordinates burned into the
frame by a vehicle-mounted camera. The model raised its overlay flag on \num{6} of the
\num{9}; the human excluded \num{8} of the \num{9}, and one of those exclusions
rests on the model's flag alone --- a watermark from the same camera model as an
image the human had already pulled, which the model matched and the human had
not yet reached. The model's flag also fires twice on overlays that name no
place at all (an \texttt{Uber} interface label, a \texttt{BLACKVUE} device
string), so \num{8} flags cover \num{6} real leaks. The one coordinate-bearing
image both reviewers passed was never sampled into the evaluation, and no image
carrying coordinates appears in it. We therefore treat neither audit as
authoritative: exclusion runs off the human verdict, with the model's leakage
flags reviewed case by case, and both columns are released so a reader can
recompute this agreement.

\subsection{The context ladder}
\label{sec:ladder}

For each site we retrieve Wikidata entities with coordinates within
\SI{6}{\kilo\metre} (\num{156}--\num{237} per site) and pass each to a
\textbf{text-only} auditor (\texttt{deepseek-v4-pro}) that returns:
\emph{(i)} usability as a reference point, \emph{(ii)} referenceability on
$0$--$5$, \emph{(iii)} a canonical English name. The prompt is reproduced in
Appendix~\ref{app:prompt-ladder}.

\paragraph{Referenceability is not fame.} The prompt states this explicitly:
the quantity is ``how often would locals or tourists use it to describe where
they are'', not ``how well known is it''. This distinction is empirically
necessary. Wikipedia pageviews rank San Jose's Rosicrucian Park (\num{5846}
views) above the Municipal Rose Garden (\num{2294}), but the former is a
curiosity museum nobody navigates by. \emph{Pageviews measure how many people
look a place up, not how many people use it to say where they are.}

\paragraph{Three construction constraints.}
\begin{enumerate}
  \item \textbf{The auditor must not be an evaluated model}, else the benchmark
  sets its own exam. Our auditor is text-only, so it cannot see the images at
  all.
  \item \textbf{Every audit decision is logged} and released
  (\texttt{audit\_decisions.jsonl}), or the benchmark is not reproducible.
  \item \textbf{Disagreement between the LLM and pageviews becomes a human
  review queue}, not an automatic override.
\end{enumerate}

\paragraph{Auditor robustness.} Cohen's $\kappa = 0.919$ across auditor models
and $0.906$ across prompt languages on the referenceability tier assignment.

\paragraph{Exclusion of contentious names.} A separate audit pass removes place
names tied to Nazism/fascism, Japanese militarism, or the celebration of
slavery, which we keep distinct from the usability field because
referenceability is a measurement while exclusion is a dataset policy. It is
deliberately narrow and flags \num{17} of \num{8676} candidates; all decisions
ship in \texttt{audit\_sensitive.jsonl}.

Stratifying by distance ($0.5$--$1.5$, $1.5$--$3$, $3$--$6$\,\si{\kilo\metre})
$\times$ referenceability tier (high $4$--$5$, mid $2$--$3$, low $0$--$1$) with
\num{2} draws per cell gives up to \num{18} reference points per site. Cells run
short in practice (Sec.~\ref{sec:limits}), so per-site counts must be weighted
in analysis. Draws within a cell are \emph{random}, not sorted by distance to
the band center: sorting cost us the single most diagnostic condition in an
early version, where a referenceability-5 landmark lost to a
referenceability-4 metro station by \SI{0.02}{\kilo\metre}.

\section{Task 1: \textsc{GeoHint}}
\label{sec:geohint}

\textsc{GeoHint} asks the model where a photograph was taken, given a
location hint that is true but coarse. The manipulation is entirely in the
hint: the image is held fixed while the reference point moves along the two
axes of Sec.~\ref{sec:ladder}.

\subsection{Task definition and prompts}

All conditions share one output schema and differ only in a prefix sentence.

\begin{itemize}
  \item \textbf{Baseline} (\texttt{context=none}): ``Where was this photo taken?
  Look at the image carefully and infer.''
  \item \textbf{Context} (\texttt{forced}): prefixed with ``I took this photo
  near $X$.''
  \item \textbf{Warned} (\texttt{forced\_warn}): as above, plus ``Important:
  the location the user mentions \textbf{may be inaccurate}, or may refer to a
  different nearby area. Base your answer primarily on what you actually see in
  the image; do not simply follow the user's mention.''
\end{itemize}

A fourth condition, \texttt{forced\_chain}, appends a structured evidence
checklist and is described where it is used (Sec.~\ref{sec:chain}). All four
are reproduced in full in Appendix~\ref{app:prompt-geohint}.

The schema requires \texttt{city}, \texttt{area} and \texttt{building} with
\emph{no} ``unknown'' escape, plus separate confidence fields---the forced-choice
design standard in psychophysics. This was necessary: under the permissive
schema \SI{43.6}{\percent} of responses decline to name a building at all,
leaving the entity-level question unanswered, against \SI{1.6}{\percent} under
the forced schema. We keep the permissive schema as a released control
condition, and we do \emph{not} post-hoc filter to finely-answered rows, since
models answer finely when confident and confidence tracks easiness.


\subsection{Metric}
\label{sec:scoring}

\paragraph{From binary hit to continuous distance.}

Binary scoring (``does the answer contain \emph{Shibuya}?'') is acutely
sensitive to whether a neighborhood has a usable name, so it is not comparable
across sites. It also collapses ``Lower Manhattan'' (off by
$\approx\SI{1}{\kilo\metre}$) and ``Chicago'' (off by \SI{1200}{\kilo\metre})
into the same wrong answer.

We instead geocode the model's answer and measure haversine distance \errkm{} to
the site's ground-truth coordinate, trying \texttt{building} $\to$ \texttt{area} $\to$
\texttt{city} and recording which level resolved. Ambiguous names are
disambiguated using an anchor derived from \textbf{the model's own answered
city}, never the ground truth---using the truth to disambiguate launders a wrong
answer (``London Soho'') into a right one (``New York SoHo'').

\subsection{Four faults that reversed the sign of the result}
\label{sec:faults}

Building this metric surfaced four scoring faults, every one of which was
invisible in aggregate statistics and was found only by reading individual
records. We report them because each produces plausible-looking numbers rather
than errors, and because together they had reversed the sign of our own headline
result before we found them:

\begin{enumerate}
\item \textbf{Parenthetical aliases silently degrade to city level.}
  \texttt{Shibuya} resolves under exact-label search but \texttt{Shibuya
  (Miyamasuzaka)} does not, falling back to a plausible-looking
  \SI{3.42}{\kilo\metre} rather than to an error.
\item \textbf{Linear features cannot be scored as points.}
  \texttt{place=Meiji-dōri} is correct with \texttt{area=Shibuya} and wrong with
  \texttt{area=Harajuku}; the two are indistinguishable at the \texttt{place}
  level, so the information must be read from \texttt{area}.
\item \textbf{A global fallback fabricates enormous errors.} Falling back to the
  globally most-linked entity turned an unverifiable answer into a
  \SI{16609}{\kilo\metre} error. Treating out-of-anchor as a level failure cut
  $>\SI{1000}{\kilo\metre}$ errors from \num{976} to \num{459}.
\item \textbf{The city-level fallback distance is a constant, and at one site it
  sits below the hit threshold.} Paris's city centroid is \SI{0.761}{\kilo\metre}
  from the site, so every answer resolving no finer than ``Paris'' counted as a
  sub-kilometer hit, inflating that site by \num{20.7} points and masking the
  accuracy drop reported in Sec.~\ref{sec:regime-model}.
\end{enumerate}

Together these corrections cut city-level resolution from \SI{51.0}{\percent}
to \SI{9.0}{\percent} of responses across the full release, and
\texttt{hit()} now treats city-level
resolution as \emph{censored}: it states only that the answer is no finer than
the city. Appendix~\ref{app:geocoder} explains why we do not substitute a
commercial geocoder.

\section{Task 2: \textsc{GeoVerify}}
\label{sec:geoverify}

\subsection{Why a second task}

\textsc{GeoHint} inherits a measurement chain that we have already shown to be
fragile: the model's free text must be geocoded, the geocode must resolve at a
usable granularity, and a threshold must be chosen (Sec.~\ref{sec:faults}
documents four faults in that chain, one of which reversed our headline
result). It also asks a question with no deployed decision attached to it: a
tourist who receives the wrong building name is misled, but nothing acts on the
answer.

\textsc{GeoVerify} removes both problems. It asks a binary question whose
ground truth is mechanical, and it corresponds to a decision that platforms
already want to automate: a driver claims to have completed a drop-off, the
platform asks for a photograph, and something must decide whether the
photograph is consistent with the claimed destination. Patents for exactly this
mechanism predate current vision--language models,\footnote{E.g.\ Niantic's
``verifying a player's real world location using image data of a landmark''
family (US 10549198, 10828569, 11325042, 11771996), filed to deter GPS spoofing
in location-based games.} but the proof-of-location literature that grew up
around them is built on GNSS cross-checks, radio fingerprints and inertial
sensors; we are not aware of a benchmark that asks whether the \emph{visual}
channel can carry this decision at the tolerance a real deployment needs.

\subsection{Task definition}

Let $g_{\mathrm{img}}$ be the true camera position of an image and
$g_{\mathrm{cand}}$ the coordinate of a named candidate place. Fix a tolerance
$\tau$. The model is shown the image and the candidate's \emph{name only}, and
asked

\begin{quote}
\emph{Was this photo taken within $\tau$ of ``\{candidate\}''?}
\end{quote}

and must answer \texttt{yes} or \texttt{no} together with a confidence in
$[0,1]$. The ground truth is
\begin{equation}
y \;=\; \mathbb{1}\!\left[\,d(g_{\mathrm{img}}, g_{\mathrm{cand}}) < \tau \,\right],
\label{eq:truth}
\end{equation}
where $d(\cdot,\cdot)$ is haversine distance. We set
$\tau = \SI{150}{\metre}$, the order of magnitude at which ride-hailing and
delivery platforms treat an arrival as complete. Nothing about
Eq.~\ref{eq:truth} depends on a geocoder, a judge model, or a tunable
threshold on the model's output: the candidate coordinates come from Wikidata
and the image coordinates from the imagery provider's metadata.

The prompt is reproduced in Appendix~\ref{app:prompt-geoverify}. The candidate
is presented \emph{by name alone}. We give no coordinates, no
city, and no description --- both because supplying them would hand over half
the answer, and because the deployed interface (``your destination:
\{name\}'') has only a name.

\paragraph{Trials.} For each image we draw two kinds of trial from the same
audited reference-point pool used by the context ladder
(Sec.~\ref{sec:ladder}):

\begin{itemize}
  \item \textbf{Signal trials}: candidates with
  $d(g_{\mathrm{img}}, g_{\mathrm{cand}}) < \tau$. Correct answer \texttt{yes}.
  \item \textbf{Noise trials}: candidates drawn from five decoy bands ---
  \SIrange{0.15}{0.3}{\kilo\metre}, \SIrange{0.3}{0.7}{\kilo\metre},
  \SIrange{0.7}{1.5}{\kilo\metre}, \SIrange{1.5}{3}{\kilo\metre} and
  \SIrange{3}{6}{\kilo\metre}. Correct answer \texttt{no}.
\end{itemize}

The nearest band starts immediately above $\tau$ on purpose: ``dropped off one
street away'' is both the hardest discrimination and the one a deployment
actually has to make. Decoys are matched to the signal candidate on
referenceability (within one tier) wherever the cell allows it, so that a model
cannot reject a decoy on type alone --- rejecting ``a famous museum'' because
the photograph shows no museum is a category judgement, not a spatial one.

\subsection{Metric: sensitivity separated from bias}
\label{sec:geoverify-metric}

Raw accuracy is not usable here. A model that answers \texttt{no} to everything
scores perfectly on every noise trial while having no spatial discrimination at
all, and the ratio of signal to noise trials is a property of our sampling
rather than of the model. We therefore score with signal detection
theory~\citep{green1966,macmillan2005}, which separates \emph{how well the
model can tell the two cases apart} from \emph{how willing it is to say yes}.

Write $H$ for the hit rate and $F(\delta)$ for the false-alarm rate at decoy
distance $\delta$:
\begin{equation}
H = P(\texttt{yes} \mid \text{signal}),
\qquad
F(\delta) = P(\texttt{yes} \mid \text{decoy at distance } \delta).
\end{equation}
In other vocabularies $H$ is the true-positive rate or recall, and $F$ the
false-positive rate; in a deployment that screens location claims, $F$ is the
rate at which a false claim is accepted, and we call it the false-acceptance
rate when discussing that reading.

Under the equal-variance Gaussian model, sensitivity and criterion are
\begin{equation}
d'(\delta) \;=\; z(H) - z\!\left(F(\delta)\right),
\qquad
c(\delta) \;=\; -\tfrac{1}{2}\left[\,z(H) + z\!\left(F(\delta)\right)\right],
\label{eq:dprime}
\end{equation}
with $z(\cdot)$ the standard normal quantile function. $d' = 0$ means the model
is exactly as likely to say \texttt{yes} when the candidate is
\SI{200}{\metre} away as when it is the right place --- no discrimination,
whatever the raw accuracy happens to be. $c < 0$ marks a liberal criterion
(biased toward confirming), $c > 0$ a conservative one. Because $z$ diverges at
$0$ and $1$, we apply the standard log-linear correction~\citep{hautus1995},
adding $0.5$ to each cell count and $1$ to each total before computing rates;
with the cell sizes reported below this bounds $|d'|$ well above any value we
observe, so no estimate is at the correction's ceiling.

Reporting $d'$ from Eq.~\ref{eq:dprime} as a function of $\delta$ turns the task into a psychometric
curve rather than a single score. The summary we propose for leaderboard use is
the \textbf{discrimination radius}
\begin{equation}
\delta_{1} \;=\; \min\left\{\, \delta : d'(\delta) \ge 1 \,\right\},
\label{eq:delta1}
\end{equation}
the decoy distance at which a model first reaches conventional discriminability.
It is a single number in meters with a directly operational reading: below
$\delta_1$, this model cannot be relied on to tell a true drop-off from a false
one.

\paragraph{Confidence.} Because the deployed failure is not ``wrong'' but
``confidently wrong'', we additionally report the confidence distribution of
false alarms. A verifier that fails but reports low confidence can be gated; one
that fails at the same confidence with which it succeeds cannot.

\subsection{Controls}

Two control arms run alongside the main arm on the same candidates.

\begin{description}
  \item[Mismatched image.] The candidate is held fixed and the image is replaced
  by street-level imagery from a different continent. Every trial is a noise
  trial and $F$ should be near zero. If it is not, the model is reasoning from
  the name rather than the photograph and the task does not measure what it
  claims to.
  \item[Blank image.] The image is replaced by a uniform gray field, holding the
  request shape constant while removing all visual evidence. This bounds what
  can be achieved from the name alone.
\end{description}

\paragraph{Pre-registered read-out.} Fixed before running, and reproduced here
verbatim from the analysis script: the task is invalid if mismatched-image
$F > \SI{20}{\percent}$, or if blank-image $F$ comes within \num{5} points of
the main arm; the benchmark is uninformative (models already solve it) if
$d' > 2$ in the nearest band; and it is informative if $d' < 0.5$ in the
nearest band while $d' > 1.5$ in the farthest.

\section{Results}

\num{5} models (\texttt{gemini-3.6-flash}, \texttt{claude-opus-5},
\texttt{claude-haiku-4-5}, \texttt{qwen3-vl-235b}, \texttt{qwen3-vl-8b}),
\num{21933} scored responses across \num{109} sites in \num{30} cities, English
only. \texttt{claude-opus-5} is sampled at one third the rate of the others on
the automatic batch, its per-call cost being an order of magnitude higher; all
comparisons involving it are either within-model or strictly paired, so the
lower rate widens its intervals without biasing any contrast.

Every $p$-value below treats the \emph{site} as the unit of resampling, not the
individual response. Images at one site show the same buildings and the same
street, and the reference points offered there come from one pool, so responses
within a site are correlated and counting each as an independent draw would
overstate the evidence. Regressions use cluster-robust standard errors grouped
by site; paired tests flip the sign of a whole site at a time rather than of
one item at a time. Point estimates are unaffected --- every percentage, slope
and ratio reported here is the same either way --- and the clustered standard
errors run \numrange{0.9}{1.8} times the unclustered ones.

\label{sec:results-hint}

\subsection{Task performance}
\label{sec:task-perf}

This section reports how well the task is completed --- whether the model names
a place close to where the photograph was taken. Sec.~\ref{sec:reliance} then
turns to a separate question: how much the model leans on the hint it was given.

\subsubsection{Aggregate accuracy}
\label{sec:aggregate}

Table~\ref{tab:models} reports accuracy and echo rate per model. A ``hit'' is
$\errkm<\SI{1}{\kilo\metre}$ \emph{and} resolution finer than city level.

\begin{table}[!htbp]
\centering
\caption{Per-model results, \texttt{forced} schema, context conditions only
(baseline rows excluded). Bands are the distance from the true location to the
reference point the model was handed. Echo rate is the fraction of responses
naming that reference point. \texttt{claude-opus-5} has fewer trials because it
was sampled at one third the rate of the others, its per-call cost being an
order of magnitude higher.}
\label{tab:models}
\begin{tabular}{lrrrrrr}
\toprule
& & & \multicolumn{3}{c}{Hit rate by band (\si{\kilo\metre})} & \\
\cmidrule(lr){4-6}
Model & $n$ & Hit rate & $0.5$--$1.5$ & $1.5$--$3$ & $3$--$6$ & Echo rate \\
\midrule
\texttt{claude-opus-5} & 1208 & \SI{52.6}{\percent} & \SI{62.0}{\percent} & \SI{49.5}{\percent} & \SI{47.9}{\percent} & \SI{19.2}{\percent} \\
\texttt{gemini-flash} & 1788 & \SI{51.2}{\percent} & \SI{63.4}{\percent} & \SI{47.8}{\percent} & \SI{44.1}{\percent} & \SI{12.2}{\percent} \\
\texttt{qwen3-vl-235b} & 1783 & \SI{32.6}{\percent} & \SI{46.0}{\percent} & \SI{29.2}{\percent} & \SI{24.6}{\percent} & \SI{24.6}{\percent} \\
\texttt{qwen3-vl-8b} & 1790 & \SI{30.6}{\percent} & \SI{46.0}{\percent} & \SI{23.1}{\percent} & \SI{24.5}{\percent} & \SI{42.3}{\percent} \\
\texttt{claude-haiku-4-5} & 1773 & \SI{25.0}{\percent} & \SI{39.8}{\percent} & \SI{20.6}{\percent} & \SI{16.7}{\percent} & \SI{22.8}{\percent} \\
\bottomrule
\end{tabular}
\end{table}

Three observations. First, the models fall into three accuracy groups:
\texttt{claude-opus-5} and \texttt{gemini-flash} are close to level at
\SI{52.6}{\percent} and \SI{51.2}{\percent}, the two \texttt{qwen3-vl} models
follow at \SI{32.6}{\percent} and \SI{30.6}{\percent} --- the \num{235}B and
\num{8}B versions differ by \num{2.0} points --- and
\texttt{claude-haiku-4-5} is lowest at \SI{25.0}{\percent}.

Second, hit rate is highest in the nearest band for every model. Part of this
follows from the scoring threshold: a hit requires an error below
\SI{1}{\kilo\metre}, and in the \SIrange{0.5}{1.5}{\kilo\metre} band the
reference point is itself close enough that naming it lands inside that radius a
fair share of the time. Between the two farther bands the rates are similar
(differences of \numrange{1.4}{4.6} points, in one case in the opposite
direction), so most of the variation with distance occurs in the first step
rather than accumulating across the range.

Third, echo rate runs broadly opposite to hit rate: it is
\SI{12.2}{\percent} and \SI{19.2}{\percent} for the two most accurate models and
\SI{42.3}{\percent} for \texttt{qwen3-vl-8b}, with
\texttt{claude-haiku-4-5} at \SI{22.8}{\percent} the main departure from the
ordering. With \num{5} models we report this as an observed ordering and attach
no statistic to it. Sec.~\ref{sec:regime-model} examines how echo rate relates
to image legibility and to accuracy within each model.

\subsubsection{Two regimes, separated by whether the image is legible}
\label{sec:baseline}

\begin{figure}[t]
\centering
\includegraphics[width=.72\textwidth]{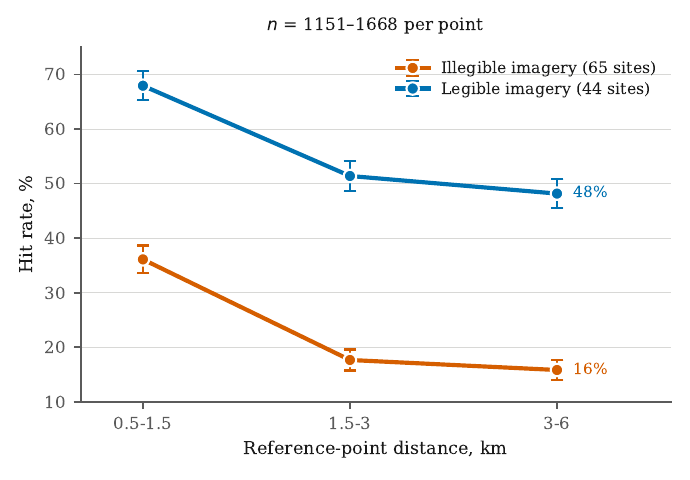}
\caption{Hit rate by reference-point distance, \texttt{forced} schema, with
sites split by whether their imagery is legible. Each plotted point rests on
\numrange{1151}{1668} observations; error bars are \SI{95}{\percent} binomial
intervals. The gap between the two curves is stable across all three distance
bands and neither curve's interval approaches the other's at any band. Both
curves are built from context conditions only, which the regime assignment
never touches; the one comparison that does read the no-context column is
corrected separately in Table~\ref{tab:splithalf}.}
\label{fig:regimes}
\end{figure}

Figure~\ref{fig:regimes} is the headline result.
The control is the \emph{no-context baseline}: same image, same model, same
schema, minus the ``I took this near $X$'' sentence. A model merely guessing
around the reference point could only gain from the context; it cannot fall
\emph{below} an uninformative baseline, so a drop below it requires a different
mechanism. That reading holds only once the baseline is estimated separately
from the split it defines, which Sec.~\ref{sec:regime-model} does.

Splitting sites at a \SI{50}{\percent} no-context hit rate gives two regimes
that differ in kind, not degree: \num{65} sites fall below the line and
\num{44} above it.

One property of this split should be stated before the results. Each site's
no-context hit rate is estimated from a median of \num{5} responses, so the
assignment of individual sites near the threshold is noisy. The analyses in
Sec.~\ref{sec:regime-model} compare each model against its own no-context
baseline on the same sites, so a site assigned to the wrong side of the
threshold moves both columns together rather than creating a difference between
them; the interventions of Sec.~\ref{sec:interventions} are paired within site
and do not use the split at all.

\subsubsection{The same split, per model}
\label{sec:regime-model}
Table~\ref{tab:regimemodel} repeats the breakdown of Table~\ref{tab:models}
within each regime. Reading the two halves against each other shows where the
pooled numbers of Sec.~\ref{sec:aggregate} come from.

\begin{table}[!htbp]
\centering
\caption{Per-model results within each regime, \texttt{forced} schema. Sites are
assigned to a regime by a fixed \SI{50}{\percent} threshold on their no-context
hit rate (\num{65} illegible, \num{44} legible). ``No context'' is the
baseline condition for the same sites and model; the remaining columns cover the
context conditions, with $n$ counting those. Rows ordered by overall accuracy.}
\label{tab:regimemodel}
\small
\begin{tabular}{lrrrrrrr}
\toprule
& & & & \multicolumn{3}{c}{Hit rate by band (\si{\kilo\metre})} & \\
\cmidrule(lr){5-7}
Model & $n$ & No context & With context & $0.5$--$1.5$ & $1.5$--$3$ & $3$--$6$ & Echo rate \\
\midrule
\multicolumn{8}{l}{\emph{Illegible sites (\num{65})}} \\
\texttt{claude-opus-5} & 649 & \SI{34.0}{\percent} & \SI{39.6}{\percent} & \SI{48.2}{\percent} & \SI{37.9}{\percent} & \SI{34.4}{\percent} & \SI{23.9}{\percent} \\
\texttt{gemini-flash} & 974 & \SI{47.2}{\percent} & \SI{34.8}{\percent} & \SI{49.5}{\percent} & \SI{30.7}{\percent} & \SI{26.4}{\percent} & \SI{15.7}{\percent} \\
\texttt{qwen3-vl-235b} & 974 & \SI{7.9}{\percent} & \SI{17.7}{\percent} & \SI{33.6}{\percent} & \SI{12.1}{\percent} & \SI{9.6}{\percent} & \SI{30.1}{\percent} \\
\texttt{qwen3-vl-8b} & 974 & \SI{7.9}{\percent} & \SI{14.2}{\percent} & \SI{29.8}{\percent} & \SI{7.4}{\percent} & \SI{7.3}{\percent} & \SI{51.1}{\percent} \\
\texttt{claude-haiku-4-5} & 974 & \SI{4.5}{\percent} & \SI{12.2}{\percent} & \SI{23.7}{\percent} & \SI{7.1}{\percent} & \SI{7.3}{\percent} & \SI{25.3}{\percent} \\
\midrule
\multicolumn{8}{l}{\emph{Legible sites (\num{44})}} \\
\texttt{claude-opus-5} & 559 & \SI{85.4}{\percent} & \SI{67.6}{\percent} & \SI{78.4}{\percent} & \SI{62.8}{\percent} & \SI{63.6}{\percent} & \SI{13.8}{\percent} \\
\texttt{gemini-flash} & 814 & \SI{91.5}{\percent} & \SI{70.8}{\percent} & \SI{79.8}{\percent} & \SI{68.4}{\percent} & \SI{65.3}{\percent} & \SI{8.1}{\percent} \\
\texttt{qwen3-vl-235b} & 809 & \SI{70.4}{\percent} & \SI{50.7}{\percent} & \SI{60.6}{\percent} & \SI{50.0}{\percent} & \SI{42.9}{\percent} & \SI{17.9}{\percent} \\
\texttt{qwen3-vl-8b} & 816 & \SI{64.3}{\percent} & \SI{50.1}{\percent} & \SI{65.1}{\percent} & \SI{41.9}{\percent} & \SI{45.1}{\percent} & \SI{31.7}{\percent} \\
\texttt{claude-haiku-4-5} & 799 & \SI{42.9}{\percent} & \SI{40.6}{\percent} & \SI{59.3}{\percent} & \SI{37.1}{\percent} & \SI{28.1}{\percent} & \SI{19.9}{\percent} \\
\bottomrule
\end{tabular}
\end{table}

Three points. First, the accuracy ordering of the models is nearly the same in
both regimes: \texttt{qwen3-vl-235b}, \texttt{qwen3-vl-8b} and
\texttt{claude-haiku-4-5} hold positions three to five in each half, while
\texttt{claude-opus-5} and \texttt{gemini-flash} exchange first and second
(\SI{39.6}{\percent} against \SI{34.8}{\percent} at illegible sites,
\SI{67.6}{\percent} against \SI{70.8}{\percent} at legible ones).

Second, echo rate is higher at illegible sites for every model, by
\numrange{5.4}{19.4} points, and the ordering across models is preserved:
\texttt{qwen3-vl-8b} echoes most in both halves (\SI{51.1}{\percent} and
\SI{31.7}{\percent}), \texttt{gemini-flash} least (\SI{15.7}{\percent} and
\SI{8.1}{\percent}).

Third, the ``No context'' column needs a correction before it can be compared
with the others. Each site's regime is assigned from that same column, so a
site that lands in the legible group partly because its baseline draw came out
high carries that upward noise into the number the context conditions are then
measured against. The comparison is biased against context at legible sites and
in favour of it at illegible ones, with no true effect required. The effect is
diluted --- each model contributes about a fifth of the pooled baseline that
assigns the regime --- but not removed.

We correct it by splitting each site's baseline responses at random, assigning
the regime from one half and reading the no-context column off the other, over
\num{200} splits (Table~\ref{tab:splithalf}). Read uncorrected, three models
are less accurate with a near reference point than without one at legible
sites. Corrected, one is: \texttt{gemini-flash}, at $-\num{8.4}$ points.
\texttt{claude-opus-5} and \texttt{qwen3-vl-235b} move to within one standard
deviation of zero, and the two weakest models gain more than the uncorrected
figures show. At illegible sites every corrected gain stays large.

Two things support reading the corrected column rather than the raw one. An
independent estimate that uses no regime split at all --- each model's
no-context rate against its near-band rate over all \num{109} sites, paired by
image --- gives $+\num{4.8}$, $-\num{3.9}$, $+\num{8.9}$, $+\num{11.0}$ and
$+\num{19.3}$ in the same model order, agreeing that
\texttt{gemini-flash} alone is worse with the hint. And the correction can only
push a group contrast toward zero, never away from it, so the gains that
survive it are conservative.

This correction is needed for one comparison and no other. Everywhere else the
regime appears --- the error ratios of Sec.~\ref{sec:ratio}, the slopes of
Sec.~\ref{sec:dist-regime}, the echo rates of Sec.~\ref{sec:echo} and the
intervention tables of Sec.~\ref{sec:interventions} --- the quantities set
against each other are context conditions, and context responses never enter
the assignment, so the split is exogenous to them. Re-deriving those quantities
from a deliberately noisier assignment, one that uses half of each site's
baseline responses, moves every one of them toward the two groups agreeing
rather than apart: the illegible near-band ratio holds at \num{1.00}, the
echo-rate gap narrows from \num{11.2} to \num{8.4} points and the ratio between
the two slopes from \num{1.56} to \num{1.41}. Assignment noise attenuates these
contrasts; it cannot manufacture them, and the full-baseline assignment the
paper uses is the less noisy of the two.

\begin{table}[!htbp]
\centering
\small
\caption{The no-context comparison before and after correcting for the fact
that the no-context column also assigns the regime. ``Uncorrected'' is the
near-band figure minus the ``No context'' column of Table~\ref{tab:regimemodel};
``corrected'' assigns the regime from a random half of each site's baseline
responses and measures the column on the other half, averaged over \num{200}
splits. An average of \num{19} of the \num{109} sites change side per split.}
\label{tab:splithalf}
\begin{tabular}{llrrr}
\toprule
Regime & Model & Uncorrected & Corrected & Shift \\
\midrule
\multirow{5}{*}{Illegible}
 & \texttt{claude-opus-5}    & $+14.2$ & $+9.3 \pm 7.1$  & $-4.8$ \\
 & \texttt{gemini-flash}     & $+2.3$  & $-3.0 \pm 4.8$  & $-5.3$ \\
 & \texttt{qwen3-vl-235b}    & $+25.7$ & $+19.6 \pm 4.9$ & $-6.1$ \\
 & \texttt{qwen3-vl-8b}      & $+22.0$ & $+16.9 \pm 4.0$ & $-5.1$ \\
 & \texttt{claude-haiku-4-5} & $+19.2$ & $+17.8 \pm 4.0$ & $-1.4$ \\
\midrule
\multirow{5}{*}{Legible}
 & \texttt{claude-opus-5}    & $-7.0$  & $+1.4 \pm 5.6$  & $+8.3$ \\
 & \texttt{gemini-flash}     & $-11.7$ & $\mathbf{-8.4 \pm 4.4}$ & $+3.3$ \\
 & \texttt{qwen3-vl-235b}    & $-9.8$  & $+2.6 \pm 5.5$  & $+12.3$ \\
 & \texttt{qwen3-vl-8b}      & $+0.8$  & $+10.6 \pm 5.6$ & $+9.8$ \\
 & \texttt{claude-haiku-4-5} & $+16.4$ & $+20.2 \pm 5.2$ & $+3.8$ \\
\bottomrule
\end{tabular}
\end{table}

\subsection{Reliance on the hint}
\label{sec:reliance}

The measures above ask whether the answer was right. The three that follow ask
where the answer came from: how far the model moved with the reference point it
was handed. Each is defined only for the context conditions, and each is
reported against the regime split of Sec.~\ref{sec:baseline}.

\subsubsection{Echo rate}
\label{sec:echo}

The \emph{echo rate} is the fraction of responses whose \texttt{area} or
\texttt{building} field names the reference point itself.

\paragraph{Echo rate by regime.} Models name the reference point in their
answer more often at illegible sites: \SI{29.6}{\percent} of contexted responses
($1345/4545$), against \SI{18.6}{\percent} ($706/3797$) at legible ones. The
per-model rates are in the last column of Table~\ref{tab:regimemodel}. Every
model echoes more in the illegible half, and the ordering across models is the
same in both halves.

One reading of this pattern is that echoing is what a model produces when it
cannot identify the scene and is required to answer anyway: the schema forbids
``unknown'' (Sec.~\ref{sec:geohint}), so some place name has to be produced, and
the reference point is the one name the prompt supplies. Echoing was not
manipulated in the design, so this is an interpretation of the pattern rather
than a result the experiment establishes.

\label{sec:mechanism}
\begin{figure}[t]
\centering
\includegraphics[width=.78\textwidth]{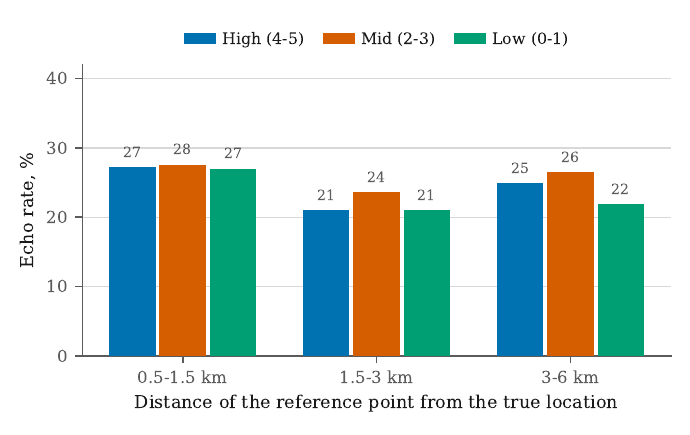}
\caption{Echo rate by reference-point distance and referenceability tier,
\texttt{forced} schema, \num{109} sites. Within every distance band the three
referenceability tiers agree to within about \num{5} points: a reference point
almost nobody uses for navigation is repeated about as often as one everybody
uses. Distance does have an effect, but a small and non-monotone one --- pooled
over tiers the rate falls from \SI{27.4}{\percent} to \SI{22.1}{\percent}
between the first and second band and returns to \SI{24.6}{\percent} in the
third, so moving the reference point from \SI{0.5}{\kilo\metre} to
\SI{6}{\kilo\metre} away changes it by \num{2.8} points.}
\label{fig:echo}
\end{figure}

\begin{table}[!htbp]
\centering
\caption{Echo rate (\si{\percent}). Flat along both axes.}
\label{tab:echo}
\begin{tabular}{lrrrr}
\toprule
 & $0.5$--$1.5$ & $1.5$--$3$ & $3$--$6$ & All \\
\midrule
\multicolumn{5}{l}{\emph{By referenceability tier}}\\
High ($4$--$5$) & 27.3 & 21.0 & 24.9 & \textbf{24.4} \\
Mid  ($2$--$3$) & 27.6 & 23.6 & 26.5 & \textbf{25.8} \\
Low  ($0$--$1$) & 27.0 & 21.1 & 21.9 & \textbf{22.9} \\
\midrule
\multicolumn{5}{l}{\emph{Pooled}}\\
All & 27.4 & 22.1 & 24.6 & 24.6 \\
\bottomrule
\end{tabular}
\end{table}

Table~\ref{tab:echo} and Figure~\ref{fig:echo} give the core mechanistic
result, and the two axes behave differently. Referenceability does essentially
nothing: pooled over distance, a reference point that almost nobody uses for
navigation is echoed \SI{22.9}{\percent} of the time against
\SI{24.4}{\percent} for one everybody uses --- a gap of \num{1.5} points, with
the middle tier the highest of the three. This axis is one the model could act
on: how commonly a place is used to give directions is judgeable from its name,
which is the only thing the prompt supplies about it. Distance moves the rate a
little more and still not much: a change from \SI{0.5}{\kilo\metre} to
\SI{6}{\kilo\metre} takes it from \SI{27.4}{\percent} to \SI{24.6}{\percent},
and the change is not monotone --- it happens between the first and second band
and then partially reverses.

The two axes therefore support different readings. Distance is not visible in
the prompt, so a flat rate along it is the absence of a signal to act on rather
than a failure to act on one, and Sec.~\ref{sec:dist-regime} shows the models
do discount the hint where the image gives them an independent check.
Referenceability is visible, and is ignored anyway. What the two share is that
the error repeating the hint produces grows with its distance
(Sec.~\ref{sec:ratio}) while the rate of repeating it does not.

\subsubsection{Error relative to hint distance}
\label{sec:ratio}

\begin{table}[!htbp]
\centering
\caption{Distance to the reference point and distance to the model's answer, by
regime and band. The $\refkm$ and $\errkm$ columns are geometric means; the last
column is the median of the per-response ratio $\errkm/\refkm$. Computed only on
responses resolving finer than city level, so the constant city-centroid
fallback (Sec.~\ref{sec:faults}) cannot drive the ratio.}
\label{tab:ratio}
\begin{tabular}{llrrrr}
\toprule
Regime & Band (\si{\kilo\metre}) & $n$ & $\refkm$ & $\errkm$ & Median ratio \\
\midrule
\multirow{3}{*}{Illegible (65 sites)}
 & $0.5$--$1.5$ & 1163 & 1.01 & 1.27 & \textbf{1.00} \\
 & $1.5$--$3$   & 1325 & 2.22 & 2.17 & \textbf{1.00} \\
 & $3$--$6$     & 1452 & 4.20 & 3.04 & 0.90 \\
\midrule
\multirow{3}{*}{Legible (44 sites)}
 & $0.5$--$1.5$ & 1102 & 0.95 & 0.54 & 0.69 \\
 & $1.5$--$3$   & 1194 & 2.11 & 0.75 & 0.44 \\
 & $3$--$6$     & 1323 & 4.41 & 0.99 & 0.24 \\
\bottomrule
\end{tabular}
\end{table}

Table~\ref{tab:ratio} is computed entirely within contexted responses and is
therefore separated from the baseline used to assign the regime; per-model
comparisons against the baseline appear in Sec.~\ref{sec:regime-model}, where
each model is reported separately.

Two distances are involved, and both are measured from the site's ground-truth
coordinate, which is fixed per site and shared by its images. $\refkm$ is the
distance from that coordinate to the reference point given as the hint in the
prompt, which is the quantity the distance bands control. $\errkm$ is the
distance from that coordinate to the place the model predicted. Their ratio says how far the answer landed relative to how far away
the hint was: a ratio of $1$ means the answer sits as far from the site as the
hint does, which is what naming the hint itself produces.

In the illegible regime the median ratio is $1.00$ to two decimal places in both
the near and mid bands. The model is not answering badly; it is answering with
the reference point. In the legible regime it is $0.24$--$0.69$: the image is
doing real work.

\subsubsection{Distance slope}
\label{sec:slope}
\label{sec:distance}
Sec.~\ref{sec:ratio} summarised the same two distances as a median ratio, one
value per band. Fitting them instead gives a rate: how much the error grows per
kilometre the reference point is moved away, using every response rather than a
per-band summary.

Regressing $\log_{10}\errkm$ on \refkm{} within contexted conditions gives a
positive coefficient everywhere we look: the farther away the reference point,
the worse the answer. Pooled over all \num{109} sites,
$\beta = +\num{0.0820}$ ($p = 1.5\times10^{-20}$, $n=\num{8274}$ responses
over \num{109} sites), and restricting to responses that resolved finer than
city level---so that the constant city-centroid fallback cannot drive the
slope---moves it to $+\num{0.0840}$ ($p = 3.6\times10^{-21}$). Table~\ref{tab:dist} shows the slope holds separately for
every model, with the strongest model showing the shallowest slope and the
weakest the steepest.

\paragraph{What the slope measures.} The coefficient has a scale with two
interpretable endpoints. A model that ignores the hint entirely produces an
error unrelated to how far away the hint is, giving $\beta = 0$. A model that
always answers with the hint produces $\errkm = \refkm$, which on this
distribution of reference-point distances fits a slope of $\num{0.176}$. The
pooled estimate of $+\num{0.0820}$ therefore sits at \SI{47}{\percent} of the
way from ignoring the hint to reciting it. Read on the original scale, moving
the reference point from \SI{0.5}{\kilo\metre} to \SI{6}{\kilo\metre} multiplies
the error by \num{2.8}.

\begin{table}[!htbp]
\centering
\caption{$\log_{10}\errkm \sim \refkm$ per model, \texttt{forced} schema,
context conditions only. ``Fine only'' excludes city-level resolutions. Every
slope is positive and significant.}
\label{tab:dist}
\begin{tabular}{lrrrr}
\toprule
Model & $n$ & $\beta$ & $p$ & $\beta$ (fine only) \\
\midrule
\texttt{claude-haiku-4-5} & 1767 & $+0.1046$ & $3.2\times10^{-12}$ & $+0.1026$ \\
\texttt{qwen3-vl-235b}    & 1766 & $+0.0961$ & $4.8\times10^{-12}$ & $+0.0966$ \\
\texttt{qwen3-vl-8b}      & 1777 & $+0.0812$ & $3.4\times10^{-12}$ & $+0.0805$ \\
\texttt{gemini-flash}     & 1758 & $+0.0713$ & $3.2\times10^{-10}$ & $+0.0795$ \\
\texttt{claude-opus-5}    & 1206 & $+0.0500$ & $5.6\times10^{-4}$  & $+0.0596$ \\
\bottomrule
\end{tabular}
\end{table}

\label{sec:dist-regime}
\begin{table}[!htbp]
\centering
\caption{$\log_{10}\errkm \sim \refkm$ fitted separately within each regime,
\texttt{forced} schema, context conditions only. A larger slope means the answer
degrades faster as the reference point is moved away.}
\label{tab:distregime}
\begin{tabular}{lrrrr}
\toprule
& \multicolumn{2}{c}{Illegible} & \multicolumn{2}{c}{Legible} \\
\cmidrule(lr){2-3}\cmidrule(lr){4-5}
Model & $n$ & $\beta$ & $n$ & $\beta$ \\
\midrule
\texttt{claude-opus-5}    & 647 & $+0.0684$ & 559 & $+0.0347$ \\
\texttt{gemini-flash}     & 955 & $+0.0935$ & 803 & $+0.0497$ \\
\texttt{qwen3-vl-235b}    & 970 & $+0.1223$ & 796 & $+0.0704$ \\
\texttt{qwen3-vl-8b}      & 968 & $+0.1116$ & 809 & $+0.0521$ \\
\texttt{claude-haiku-4-5} & 972 & $+0.1015$ & 795 & $+0.1110$ \\
\midrule
All                       & 4512 & $+0.1008$ & 3762 & $+0.0648$ \\
\bottomrule
\end{tabular}
\end{table}

Fitting the same regression within each regime separates the two
(Table~\ref{tab:distregime}). Pooled over models the slope is $+\num{0.1008}$ at illegible sites and
$+\num{0.0648}$ at legible ones, a factor of \num{1.56}; restricting to
responses finer than city level gives $+\num{0.1078}$ and $+\num{0.0625}$. On
the scale above, where $\num{0.176}$ corresponds to always
answering with the hint, these are \SI{57}{\percent} and \SI{37}{\percent} of
the way to that endpoint: at illegible sites the answer moves with the reference
point roughly half again as much as it does at legible ones. Four of the five
models follow that pattern individually, with slopes roughly twice as steep at
illegible sites. \texttt{claude-haiku-4-5} does not: its slopes are
$+\num{0.1015}$ and $+\num{0.1110}$, the only case where the legible value is
the larger of the two, which places it at the illegible level even on sites
where its own no-context accuracy is above the threshold. All ten coefficients are positive. Nine
reach $p < 0.01$; \texttt{claude-opus-5}'s legible-regime slope, the shallowest
of the ten and the one resting on the fewest responses, reaches
$p = \num{0.067}$.

\subsection{Interventions}
\label{sec:interventions}

Two prompt-level interventions were tested against the plain \texttt{forced}
condition on strictly paired items. The first tells the model the hint may be
wrong; the second requires it to work through a visual-evidence checklist before
answering.

\subsubsection{Warning that the hint may be wrong}
\label{sec:mitigation}

Table~\ref{tab:mitigation} compares \texttt{forced} against
\texttt{forced\_warn} on \emph{strictly paired} observations --- same site, same
image, same model, same language, same reference point --- giving \num{8215}
pairs across \num{109} sites.

\begin{table}[!htbp]
\centering
\caption{Hit rate (\si{\percent}) by regime, strictly paired. $\Delta$ is the
paired difference with a two-sided permutation test, \num{10000} sign flips.}
\label{tab:mitigation}
\begin{tabular}{llrrrrl}
\toprule
Regime & Schema & Near & Mid & Far & Overall & Paired $\Delta$ \\
\midrule
\multirow{2}{*}{Illegible}
 & \texttt{forced}       & 36.1 & 17.7 & 15.8 & 22.6 & \multirow{2}{*}{$+0.46$, $p=\num{0.49}$} \\
 & \texttt{forced\_warn} & 35.4 & 18.1 & 17.3 & 23.0 & \\
\midrule
\multirow{2}{*}{Legible}
 & \texttt{forced}       & 67.9 & 51.4 & 48.2 & 55.2 & \multirow{2}{*}{$\mathbf{+3.94}$, $\mathbf{p=\num{0.0008}}$} \\
 & \texttt{forced\_warn} & 69.4 & \textbf{55.5} & \textbf{54.2} & \textbf{59.2} & \\
\midrule
\multirow{2}{*}{All sites}
 & \texttt{forced}       & 50.6 & 33.0 & 30.5 & 37.4 & \multirow{2}{*}{$+2.01$, $p=\num{0.002}$} \\
 & \texttt{forced\_warn} & 50.7 & 35.0 & 33.9 & 39.3 & \\
\bottomrule
\end{tabular}
\end{table}

The warning is demonstrably obeyed in both regimes: echo rate falls from
\SI{29.4}{\percent} to \SI{21.0}{\percent} where the image is illegible and from
\SI{18.7}{\percent} to \SI{12.6}{\percent} where it is legible. Obedience is
therefore not what separates the two regimes --- what separates them is whether
obeying helps.

It helps only where the image is legible ($+\num{3.94}$ points,
$p=\num{0.0008}$). Where it is not, the paired difference is $+\num{0.46}$
points and indistinguishable from zero ($p=\num{0.49}$). This is what the mechanism
predicts: the warning's only action is \emph{trust the text less}, which is
worth something only if there is something else to trust. Nothing that merely
discounts the context can repair the illegible regime.

Table~\ref{tab:mitigmodel} gives the paired difference per model. Every model
gains more at legible sites than at illegible ones. The largest gains belong to
\texttt{qwen3-vl-235b} ($+\num{6.54}$ points) and \texttt{claude-opus-5}
($+\num{6.17}$); \texttt{claude-haiku-4-5}, the least accurate model in
Table~\ref{tab:models}, moves by $+\num{0.38}$.

\begin{table}[!htbp]
\centering
\caption{Paired difference in hit rate (percentage points) from adding the
warning, by model. Positive means \texttt{forced\_warn} scored higher than
\texttt{forced} on the same items. Rows ordered by overall accuracy.}
\label{tab:mitigmodel}
\begin{tabular}{lrrrr}
\toprule
Model & $n$ pairs & All sites & Illegible & Legible \\
\midrule
\texttt{claude-opus-5}    & 1200 & $+3.67$ & $+1.54$ & $+6.17$ \\
\texttt{gemini-flash}     & 1754 & $+2.51$ & $+2.05$ & $+3.08$ \\
\texttt{qwen3-vl-235b}    & 1753 & $+2.97$ & $+0.10$ & $+6.54$ \\
\texttt{qwen3-vl-8b}      & 1754 & $+1.45$ & $-0.72$ & $+4.17$ \\
\texttt{claude-haiku-4-5} & 1754 & $+0.00$ & $-0.31$ & $+0.38$ \\
\bottomrule
\end{tabular}
\end{table}

\paragraph{What each condition costs in precision.} The warning makes answers
slightly coarser. On the same strictly paired items, the share of responses
resolving at building level falls from \SI{27.9}{\percent} under
\texttt{forced} to \SI{24.7}{\percent} under \texttt{forced\_warn}, a paired
difference of $-\num{3.15}$ points ($p<10^{-4}$, $n=\num{8215}$ pairs over
\num{109} sites): told the hint
may be wrong, the model retreats to naming a neighbourhood rather than a
building. The accuracy gain reported above is therefore bought at some loss of
precision.

Supplying the context itself moves answers the other way. Building-level
resolution sits above the no-context baseline in all three bands, paired by
image ($+\num{4.3}$, $+\num{4.0}$ and $+\num{2.9}$ points, at
$p=\num{0.007}$, \num{0.010} and \num{0.046}), so the coarsening belongs to the
warning rather than to the hint.

\subsubsection{Requiring a structured evidence checklist}
\label{sec:chain}

If the failure in the illegible regime were a \emph{reasoning} failure, forcing
the model to enumerate visual evidence before answering should repair it. We
tested this directly with \texttt{forced\_chain}, a schema that requires the
model to fill a checklist of visual-clue categories adapted from
GeoRC \citep{georc} --- infrastructure, vegetation, architecture, script,
terrain --- and to write \texttt{"not visible"} rather than leave a slot empty.
(We deliberately omit GeoRC's ``meta information'' category, which in our
setting would leak the answer.) The schema ran on three of the five models, so
every comparison below is restricted to those three and strictly paired: same
site, same image, same model, same language, same reference point;
\num{2538} pairs across \num{18} sites.

We fixed the read-out criteria before running. The decisive one was the ratio of
answer error to reference-point distance (Table~\ref{tab:ratio}): a ratio of
\num{1.00} means the answer \emph{is} the reference point, and a reasoning
intervention that made models read the image should pull it down.

\begin{table}[!htbp]
\centering
\caption{\texttt{forced\_chain} versus \texttt{forced}, strictly paired, three
models. Hit rate in \si{\percent}; ratio is median error/reference distance,
computed on non-city-level resolutions only.}
\label{tab:chain}
\begin{tabular}{llrrrrr}
\toprule
Regime & Schema & Near & Mid & Far & Overall & Ratio (near/mid/far) \\
\midrule
\multirow{2}{*}{Illegible}
 & \texttt{forced}        & 37.4 & 16.7 & 15.7 & 22.2 & 1.00 / 1.01 / 0.87 \\
 & \texttt{forced\_chain} & 39.8 & 14.9 & 15.5 & 22.2 & 1.00 / 1.02 / 0.87 \\
\midrule
\multirow{2}{*}{Legible}
 & \texttt{forced}        & 66.7 & 49.0 & 41.5 & 51.2 & 0.60 / 0.50 / 0.50 \\
 & \texttt{forced\_chain} & 63.2 & 51.6 & 44.4 & 52.2 & 0.60 / 0.43 / 0.35 \\
\bottomrule
\end{tabular}
\end{table}

Every pre-registered criterion fails (Table~\ref{tab:chain}). The paired change
in hit rate is $\num{0.00}$ points in the illegible regime ($n=\num{1197}$,
$p=\num{1.00}$) and $+\num{1.04}$ in the legible one ($n=\num{1341}$,
$p=\num{0.38}$); the smallest $p$ for any individual model is \num{0.17}. The
ratio holds at \num{1.00} in the nearest band of the illegible regime and
within \num{0.02} of it in the middle band. The echo rate holds as well
(\SI{22.1}{\percent} to \SI{22.1}{\percent} illegible, \SI{15.9}{\percent} to
\SI{14.8}{\percent} legible), so the schema leaves even the wording of the
answers where it was.
For comparison, the one-sentence warning of Section~\ref{sec:mitigation} buys
$+\num{3.94}$ points in the legible regime at a fraction of the token cost.

\paragraph{The models filled the checklist in.} A checklist that no one follows
would produce the same null, so we read the raw responses back
(Table~\ref{tab:chaincomply}). An \texttt{evidence} object is present in at
least \SI{99.7}{\percent} of responses from every model; \num{2} responses out
of \num{2799} are unparseable. The three counts in this section cover different
things: \num{2799} is every \texttt{forced\_chain} response, \num{2538} of them
pair against a \texttt{forced} counterpart in a distance band, and the slot
count below runs over the \num{2797} that parsed. Of the \num{13961} category
slots,
\SI{7.0}{\percent} are answered \texttt{"not visible"}, and the rate depends on
the category: vegetation is declined between \SI{21}{\percent} and
\SI{30}{\percent} of the time, architecture never, by any model. The models use
the exit where the image gives them nothing, and architecture is a category they
can always say something about. The null result therefore describes the
intervention, under compliance with it.

\begin{table}[!htbp]
\centering
\small
\caption{\texttt{forced\_chain} compliance, \num{2799} responses. The second
column gives the share of responses carrying an \texttt{evidence} object; the
remaining columns give the share of each category's slots answered
\texttt{"not visible"}, the response the prompt requires in place of a guess.}
\label{tab:chaincomply}
\begin{tabular}{lrrrrrr}
\toprule
Model & \texttt{evidence} & Infra. & Veg. & Arch. & Script & Terrain \\
\midrule
\texttt{gemini-flash}     & \SI{99.7}{\percent}  & \SI{0.0}{\percent} & \SI{21.2}{\percent} & \SI{0.0}{\percent} & \SI{2.6}{\percent} & \SI{0.0}{\percent} \\
\texttt{qwen3-vl-235b}    & \SI{100.0}{\percent} & \SI{3.8}{\percent} & \SI{29.8}{\percent} & \SI{0.0}{\percent} & \SI{7.2}{\percent} & \SI{4.5}{\percent} \\
\texttt{claude-haiku-4-5} & \SI{99.8}{\percent}  & \SI{0.0}{\percent} & \SI{27.6}{\percent} & \SI{0.0}{\percent} & \SI{7.2}{\percent} & \SI{0.4}{\percent} \\
\bottomrule
\end{tabular}
\end{table}

This replicates, under a stricter design, EarthWhere's finding that deeper
reasoning does not reliably help when visual clues are limited
\citep{earthwhere}; their comparison was across conditions, ours is paired
within item and model. We read it as evidence about \emph{where} the deficit
sits: a checklist can only reorganise evidence the model has already extracted,
and in the illegible regime there is none to reorganise.

\subsection{\textsc{GeoVerify}: discrimination as a function of decoy distance}
\label{sec:results-verify}

We report the algorithmically selected batch only: \num{90} sites in \num{30}
cities, \num{100} images, \num{1448} trials, \num{6270} scored responses across
the same \num{5} models. Every returned response parsed; one call failed at
the API and is excluded.

\paragraph{Controls.} Both pre-registered validity checks pass overall
(Table~\ref{tab:verify}). Replacing the image with one from another
continent while keeping the claim drops false acceptances to \SI{5.0}{\percent},
against \SI{37.7}{\percent} in the main arm, and a uniform gray field gives
\SI{0.0}{\percent} for every model. The models are reading the photograph, not
the place name.

The per-model column shows what the pooled control hides.
\texttt{claude-haiku-4-5} accepts \SI{18.1}{\percent} ($36/199$) of the
swapped-continent claims --- approaching the \SI{20}{\percent} threshold at
which we had pre-committed to declaring the task invalid --- while the other
four models sit between \SI{0.0}{\percent} and \SI{3.5}{\percent}. Together with
its $d'$ below \num{1} in every band, this indicates that this model is largely
not using the image for this task.

\begin{table}[!htbp]
\centering
\small
\caption{\textsc{GeoVerify} results, one column per model. $H$ is the hit rate
on signal trials (true-positive rate) and $F$ the false-acceptance rate; the
control rows give $F$ when the image is swapped for one from another continent
and when it is replaced by a uniform gray field. $d'$ is sensitivity by decoy
distance and $\delta_1$ the discrimination radius, the nearest band reaching
$d' \geq 1$. The criterion $c$ is negative when a model is biased toward
accepting a claim and positive when biased toward rejecting one. The last two
rows describe the confidence attached to false acceptances. Log-linear
correction applied \citep{hautus1995}.}
\label{tab:verify}
\begin{tabular}{lrrrrrr}
\toprule
& \texttt{gemini-} & \texttt{claude-} & \texttt{qwen3-} & \texttt{qwen3-} & \texttt{claude-} & \\
& \texttt{flash} & \texttt{opus-5} & \texttt{vl-8b} & \texttt{vl-235b} & \texttt{haiku-4-5} & Pooled \\
\midrule
\multicolumn{7}{l}{\emph{Response rates}} \\
$H$ (signal accepted)      & \SI{93.5}{\percent} & \SI{47.2}{\percent} & \SI{44.4}{\percent} & \SI{70.4}{\percent} & \SI{78.7}{\percent} & \SI{69.9}{\percent} \\
$F$ (decoy accepted)       & \SI{35.1}{\percent} & \SI{13.1}{\percent} & \SI{18.8}{\percent} & \SI{43.8}{\percent} & \SI{61.1}{\percent} & \SI{37.7}{\percent} \\
\quad control: swapped image & \SI{3.5}{\percent} & \SI{0.0}{\percent} & \SI{0.0}{\percent} & \SI{0.0}{\percent} & \textbf{\SI{18.1}{\percent}} & \SI{5.0}{\percent} \\
\quad control: gray field  & \SI{0.0}{\percent} & \SI{0.0}{\percent} & \SI{0.0}{\percent} & \SI{0.0}{\percent} & \SI{0.0}{\percent} & \SI{0.0}{\percent} \\
\midrule
\multicolumn{7}{l}{\emph{Sensitivity $d'$ by decoy distance (\si{\kilo\metre})}} \\
$0.15$--$0.3$              & 0.68 & 0.32 & 0.51 & 0.24 & 0.22 & 0.33 \\
$0.3$--$0.7$               & \textbf{1.61} & 0.75 & 0.55 & 0.46 & 0.40 & 0.65 \\
$0.7$--$1.5$               & 2.16 & \textbf{1.24} & 0.66 & 0.68 & 0.44 & 0.87 \\
$1.5$--$3$                 & 2.34 & 1.44 & 0.89 & 0.87 & 0.55 & \textbf{1.03} \\
$3$--$6$                   & 2.36 & 1.95 & \textbf{1.13} & \textbf{1.07} & 0.82 & 1.20 \\
$\delta_1$                 & $0.3$--$0.7$ & $0.7$--$1.5$ & $3$--$6$ & $3$--$6$ & none & $1.5$--$3$ \\
\midrule
\multicolumn{7}{l}{\emph{Criterion $c$}} \\
nearest band               & $-1.14$ & $+0.23$ & $+0.39$ & $-0.41$ & $-0.68$ & $-0.36$ \\
farthest band              & $-0.31$ & $+1.04$ & $+0.70$ & $+0.00$ & $-0.37$ & $+0.08$ \\
\midrule
\multicolumn{7}{l}{\emph{Confidence on false acceptances}} \\
median                     & 0.95 & 0.60 & 0.95 & 0.95 & 0.85 & 0.85 \\
share $\geq 0.8$           & \SI{97.6}{\percent} & \SI{9.8}{\percent} & \SI{100.0}{\percent} & \SI{99.5}{\percent} & \SI{65.0}{\percent} & \SI{83.8}{\percent} \\
\bottomrule
\end{tabular}
\end{table}

\begin{figure}[!htbp]
\centering
\includegraphics[width=.82\textwidth]{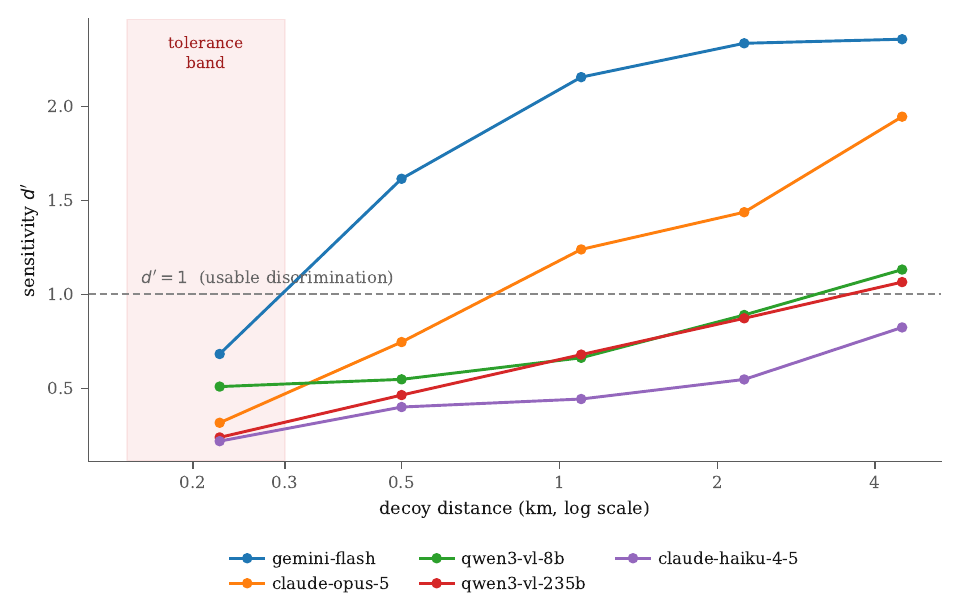}
\caption{Sensitivity against decoy distance, one line per model; the same
numbers as the $d'$ block of Table~\ref{tab:verify}, drawn to show the shape.
The dashed line marks $d'=1$, conventional usable discrimination. The shaded
strip is the band that begins immediately above the \SI{150}{\metre} tolerance
--- the discrimination a deployment actually has to make --- and no model
crosses the dashed line inside it. Every model improves as the decoy is moved
away, and they separate by a factor of three by the farthest band, so the
difficulty is a property of the distance rather than of the task as a whole.}
\label{fig:verify}
\end{figure}

\paragraph{Discrimination rises with decoy distance, but not where it matters.}
Pooled $d'$ climbs monotonically from \num{0.33} in the nearest band to
\num{1.20} in the farthest (Figure~\ref{fig:verify}), and the criterion $c$
moves from $-0.36$ to $+0.08$
over the same range. The band that matters for the deployed decision is the
nearest one --- a vehicle left one street away rather than at the requested
door --- and there no model reaches $d'=1$; the best, \texttt{gemini-flash},
manages \num{0.68}. The task is hardest exactly where a platform would need it
to work.

\paragraph{Accuracy alone would invert the ranking.} \texttt{claude-opus-5}
answers ``yes'' to only \SI{47.2}{\percent} of true claims, the second-worst hit
rate in the table, yet has the second-best sensitivity ($d'=\num{1.95}$ in the
farthest band). It is not worse at the task; it is running a far stricter
criterion: its $c$ runs from $+0.23$ to $+1.04$ across the bands, against
$-1.14$ to $-0.31$ for \texttt{gemini-flash}, where positive values mean a bias
toward rejecting a claim and negative values a bias toward accepting one. Any metric that does not separate sensitivity from bias
would rank these two models the wrong way round --- which is the empirical case
for scoring this task with signal detection theory rather than accuracy.

\paragraph{Confidence does not identify the errors.} Table~\ref{tab:verify}
gives the self-reported confidence attached to false acceptances. Pooled, their
median confidence is \num{0.85} against \num{0.95} for correct acceptances, and
\SI{83.8}{\percent} of them are stated at confidence $\geq \num{0.8}$. The
per-model spread is an order of magnitude: \texttt{claude-opus-5} states
\SI{9.8}{\percent} of its false acceptances at that confidence, while
\texttt{qwen3-vl-8b} states \SI{100.0}{\percent} of its \num{177}. A platform
cannot threshold on model-reported confidence to filter these mistakes out,
except with \texttt{claude-opus-5}.
\paragraph{Read-out against the pre-registered criteria.} We pre-committed to
calling the result informative if the nearest band gave $d' < \num{0.5}$ and the
farthest exceeded $d' > \num{1.5}$. Pooled, the first holds (\num{0.33}) and the
second does not (\num{1.20}), so by that rule the pooled result is
\emph{inconclusive}. We report this rather than adjusting the threshold after
the fact. The per-model table shows why. One model meets both halves of the
criterion: \texttt{claude-opus-5}, at $0.32$ in the nearest band and $1.95$ in
the farthest. \texttt{gemini-flash} clears the far end by the widest margin of
any model ($2.36$) and still misses the near one ($0.68 > 0.5$), which is the
result stated plainly --- the model with the best discrimination overall cannot
make the discrimination the tolerance demands. The remaining three miss the far
end, and they are what pulls the pooled curve below it. The correct summary is that model-to-model
variation on this task is large, not that the task is uniformly too hard.

\section{Limitations}
\label{sec:limits}

The benchmark inherits the geography of its imagery source. Mapillary is
volunteer-contributed and its coverage is uneven, dense across Western Europe
and North America and thin where street-level capture is regulated or the
contributor community is small, with mainland China the largest such gap. Every
candidate city we tried cleared the coverage check, but that says more about how
the candidate list was assembled --- major tourist destinations --- than about
global coverage; one city measured outside that list offered \num{141} images
that were entirely equirectangular panoramas, with no usable perspective frames.
Wikidata introduces a second bias, toward encyclopedically notable reference
points: \SI{8.1}{\percent} of the candidates it returns carry no English title,
and a place with no English-language coverage is unlikely to be returned at
all. All
prompts are in English, and one site's imagery is three to five years older than
the rest, which will have changed shopfront signage but not building-level
judgements. Whether the asymmetry we report holds for a user writing in another
language, or in a car-oriented city, is untested.

Two measurement caveats bound how far the results can be pushed. Each site's
no-context hit rate rests on a median of \num{5} responses, so sites near the
\SI{50}{\percent} threshold are assigned to a regime with real uncertainty; two
more images per site would roughly triple that baseline at a cost of a few
hundred calls, and until then the per-model results of
Sec.~\ref{sec:regime-model} are the more dependable cut. Sparse ladder cells,
notably low-referenceability at far distances, yield fewer reference points than
the design allows, so per-site counts must be weighted. Finally, two constructs
are narrower than their names suggest: \Ldist{} measures how finely
neighbourhoods are cut rather than how far apart they feel to someone in a taxi,
and linear features such as a long avenue resolve only to neighbourhood level,
costing precision --- \texttt{Meiji-dōri + Shibuya} scores
\SI{0.53}{\kilo\metre} against a true \SI{0.05}{\kilo\metre} --- without
affecting whether an answer counts as correct.

\section{Conclusion}

A VLM's reliance on user-supplied location context does not scale with how
informative that context is: neither the reference point's distance nor its
referenceability changes how often the model repeats it, while the error that
repeating it produces grows steadily with distance. The consequence is
conditional on the image. Where it is unreadable the answer collapses onto the
hint --- the median error equals the hint's own distance --- and the hint is the
only signal the model has. Where it is readable the image does real work, and
one model of the five is measurably worse for having been handed true
information --- a smaller claim than the raw regime comparison suggests, and
the one that survives correcting that comparison for its own selection bias.
Telling the model the user may be wrong is obeyed in both cases but only pays
where there is something else to trust; a structured evidence checklist pays
nowhere.

On the same images, verifying a location \emph{claim} is not yet reliable at the
tolerance a deployment would need: no model separates a true claim from a decoy
one street away, ranking by accuracy would order the models wrongly because they
differ sharply in how readily they say yes, and most false acceptances are
asserted confidently. We release the benchmark, the auto-generated context
ladders, every auditor decision, and the scoring code.

\section*{Reproducibility}

Code and data: \url{https://github.com/Yifan1209/geocontext}. We release image IDs and coordinates rather
than Mapillary pixels. \texttt{audit\_decisions.jsonl},
\texttt{audit\_sensitive.jsonl}, the \num{109} ladder CSVs,
\texttt{district\_scale.csv}, the city-screening and candidate tables, the
parsed and geocoded \textsc{GeoHint}
responses, the raw \textsc{GeoVerify} responses and the raw
\texttt{forced\_chain} responses that Table~\ref{tab:chaincomply} is computed
from are included. The raw response text for the remaining \textsc{GeoHint}
schemas is available from the authors on request.

\bibliographystyle{plainnat}
\bibliography{refs}

\appendix

\section{Prompts}
\label{app:prompts}

All prompts are reproduced from the released code. The prompt \emph{is} the
method here, so we ship it rather than paraphrase it.

\subsection{\textsc{GeoHint}}
\label{app:prompt-geohint}

The context sentence, prepended to the schema, is
\verb|I took this photo near <reference point>. |
including the trailing space. \texttt{v1} permits abstention and serves as the
control; \texttt{forced} is the schema all reported results use. Long lines are
wrapped to fit the page.

\paragraph{\texttt{v1}.}
\begin{footnotesize}
\begin{verbatim}
<context>Where was this photo taken? Look at the image carefully and infer.

Output only JSON, no other text, in this format:
{
  "country": "country",
  "city": "city",
  "area": "the specific commercial district / neighborhood / landmark area,
           or unknown",
  "place": "the most specific place or shop name, or unknown",
  "confidence": a number between 0.0 and 1.0,
  "clues": ["a concrete cue you actually see in this image supporting your
            answer", "..."]
}

In "clues", list only things you **actually see in this image**. If you cannot
see something clearly, say so.
\end{verbatim}
\end{footnotesize}

\paragraph{\texttt{forced}.}
\begin{footnotesize}
\begin{verbatim}
<context>Where was this photo taken? Look at the image carefully and infer.

Output only JSON, no other text:
{
  "city": "city",
  "area": "commercial district / neighbourhood name.
           **Give your best guess -- 'unknown' is not allowed**",
  "building": "the building or shop that is the main subject.
               **Guess -- 'unknown' is not allowed**",
  "confidence_area": 0.0-1.0, your confidence in `area`,
  "confidence_building": 0.0-1.0, your confidence in `building`,
  "clues": ["a concrete cue you actually see in this image supporting your
            answer", "..."]
}

Commit to a specific name even when unsure -- express the uncertainty in the
confidence fields, do not fall back on 'unknown'.
In "clues", list only things you **actually see in this image**.
\end{verbatim}
\end{footnotesize}

\paragraph{\texttt{forced\_warn}.} The text below is appended to
\texttt{forced}.
\begin{footnotesize}
\begin{verbatim}
Important: the location the user mentions **may be inaccurate**, or may refer
to a different nearby area. Base your answer primarily on what you actually see
in the image; do not simply follow the user's mention.
\end{verbatim}
\end{footnotesize}

\paragraph{\texttt{forced\_chain}.} Appended to \texttt{forced}. The category
ordering follows the citation frequency measured over champion-level reasoning
chains by \citet{georc}; their meta-information category is dropped because it
is specific to the GeoGuessr interface, and the language category asks for
script only, so that reading place names off signage cannot substitute for
visual inference.
\begin{footnotesize}
\begin{verbatim}
Before answering, examine the image for the following categories of geographic
evidence, in this order. For each, write what you actually observe, or
"not visible" if you cannot see it. Do not guess.

  1. Infrastructure -- utility poles, bollards, kerbs, road markings, traffic
     light mounting, signage shapes
  2. Vegetation -- species type, whether tropical / temperate / arid
  3. Architecture -- building materials, window proportions, roof style,
     facade rhythm
  4. Language -- the SCRIPT of any visible text (Latin / Cyrillic / CJK /
     Arabic ...). Report the script only, NOT what the text says.
  5. Terrain, climate, vehicles, road surface

Base your answer on what you listed. Add an "evidence" object to the JSON with
one short string per category above (use "not visible" where applicable).
\end{verbatim}
\end{footnotesize}

\subsection{\textsc{GeoVerify}}
\label{app:prompt-geoverify}

One prompt for every trial, including both controls; only the image and the
candidate name change. The candidate is given by name alone, with no
coordinates, city, or description, matching what a deployed application would
show. The wording pushes toward neither verdict, so that the criterion $c$ is
measured rather than set by the prompt.

\begin{footnotesize}
\begin{verbatim}
You are checking whether a photo was taken at a claimed location.

CLAIM: This photo was taken within 150 metres of "<candidate>".

Look at the image and decide whether the claim is true.

Output only JSON, no other text:
{
  "verdict": "yes" or "no",
  "confidence": a number between 0.0 and 1.0,
  "evidence": ["specific things you actually see in the image that led to your
               verdict"]
}

You must choose "yes" or "no". Put your uncertainty in `confidence`, not in the
verdict. In `evidence`, list only what you can actually see in this image.
\end{verbatim}
\end{footnotesize}

\subsection{Reference-point audit (\texttt{deepseek-v4-pro}, text only)}
\label{app:prompt-ladder}

This is the prompt that rates each Wikidata candidate for usability and
referenceability (Sec.~\ref{sec:ladder}). The auditor never sees an image.
The \texttt{zh} field asks for a Chinese name where one exists. It is carried
through to the released ladder files but plays no part in this paper, whose
conditions are English only; we keep it in the prompt because every audit
decision we release was produced by exactly this text, and editing the prompt
would leave those decisions unreproducible. Chinese strings in the original are
shown as \texttt{<Chinese name>} below, as this document loads no CJK fonts.

\begin{small}
\begin{verbatim}
You are selecting LOCATION REFERENCE POINTS for a visual geolocation experiment.

Scenario: a tourist photographs something, asks an AI "where is this?", and
casually adds their rough whereabouts - "I took this near <somewhere>". Judge
whether each candidate works as that <somewhere>.

For each candidate give four fields:

1. `ok`: usable or not. Typical rejects:
   - Administrative divisions (districts, subdistricts, wards, boroughs) -
     nobody says "I'm near the 3rd Municipal Subdistrict"
   - Abstract systems (a metro network, a railway line) - not a spot you can
     stand next to
   - Events and incidents - the coordinate is merely where it happened
   - Objects too small to navigate by (gravestones, plaques, individual rooms)

2. `familiarity`: **referenceability**, 0-5. This is NOT "how famous is it".
   It is "how often would locals or tourists use it to describe where they are".
   A curiosity attraction may get many encyclopedia readers yet nobody navigates
   by it - score those low.
   5 = everyone uses it to locate themselves; 3 = locals use it;
   1 = almost nobody says it.

3. `zh`: canonical Simplified Chinese name. Use null if there is no common
   Chinese name.

4. `en`: canonical English name. Use null if there is none.

Candidates (with distance from the reference point):

{items}

Output ONLY a JSON array, elements like:
{"idx":0,"ok":true,"familiarity":4,"zh":"<Chinese name>",
 "en":"San Jose Municipal Rose Garden","why":"under 12 words"}
No other text.
\end{verbatim}
\end{small}

\subsection{Image audit (\texttt{deepseek-vision})}
\label{app:prompt-image}

This is the model half of the two-track image audit (Sec.~\ref{sec:resource}).
The human reviewer works from the image alone and records a keep/exclude verdict
with a free-text reason, without seeing these fields.

\begin{small}
\begin{verbatim}
Audit this street-level photo for a geolocation benchmark.

Report on the IMAGE ITSELF, not on where you think it was taken.

1. `answer_overlay`: is there text OVERLAID on the photo (app UI, news subtitle,
   watermark, caption bar) that names a place, street, district, city or
   landmark? Text physically present in the scene - shop signs, street signs,
   banners, billboards - does NOT count. Those are legitimate visual cues.
2. `overlay_text`: any overlaid text you can read (empty string if none).
3. `scene_type`: one of "street", "indoor", "vehicle_interior", "closeup",
   "other". "street" = an outdoor street-level view a pedestrian or car would
   see. "vehicle_interior" = shot from inside a vehicle with dashboard/pillars
   visible. "closeup" = a wall, sign or object filling the frame with no street
   context.
4. `through_glass`: shot through a windscreen or window (reflections, glare,
   wipers)?
5. `tourist_infrastructure`: does the scene show visual signs of being right
   next to a major tourist attraction, EVEN IF the attraction itself is not in
   frame? Look for: souvenir/gift shops, ticket booths or entrance queues, tour
   groups or guides with flags/umbrellas, dense multilingual directional signage
   aimed at visitors. Ordinary shops, ordinary street signs, and ordinary
   pedestrians do NOT count.
6. `quality`: 0-5 overall usability for geolocation.
   5 = sharp, well framed, plenty of context; 3 = usable; 0 = unusable.
7. `usable`: false if answer_overlay is true, or scene_type is not "street",
   or quality <= 1. (`tourist_infrastructure` does NOT affect `usable` - it is
   recorded for robustness analysis, not filtered out.)

Output ONLY JSON:
{"answer_overlay": true/false, "overlay_text": "", "scene_type": "street",
 "through_glass": false, "tourist_infrastructure": false, "quality": 4,
 "usable": true, "why": "under 15 words"}
\end{verbatim}
\end{small}

\section{Why not a commercial geocoder}
\label{app:geocoder}

On the \num{100} labels Wikidata
currently fails, Nominatim resolves \num{15} (net new \num{10}). Classifying the
\num{85} both fail: \num{36} are descriptive phrases (``Friday's or an adjacent
commercial building''), \num{21} are generic categories (``roadside public
restroom''), \num{28} are plausible names---\num{8} of which are bank chains with
thousands of branches and hence not uniquely locatable. The ceiling is
$\approx\SI{3}{\percent}$ of records, against a cost of new cross-city
fuzzy-match errors. Google Places is better still, but its terms restrict
caching and dataset creation, which conflicts with releasing resolved
coordinates for reproducibility.

\end{document}